\documentclass{article}

\usepackage{times}
\usepackage{graphicx} % more modern
\usepackage{subfigure} 
\usepackage{amsmath,xspace}
\usepackage{pgfplots}
\usepackage{color}

\usepackage{natbib}

\usepackage{algorithm}
\usepackage{algorithmic}
\usepackage{amsmath}
\usepackage{amssymb}

\usepackage{hyperref}

\usepackage[accepted]{icml2018} 

\newcommand\drqa{DrQA\xspace}
\newcommand{\us}{Weaver\xspace}
\newcommand{\rrr}{Reinf. reader-ranker\xspace}
\newcommand{\rd}[1]{\textcolor{red}{#1}}

\begin{document} 

\twocolumn[
\icmltitle{\us: Deep Co-Encoding of Questions and Documents for Machine Reading}

% It is OKAY to include author information, even for blind
% submissions: the style file will automatically remove it for you
% unless you've provided the [accepted] option to the icml2017
% package.

% list of affiliations. the first argument should be a (short)
% identifier you will use later to specify author affiliations
% Academic affiliations should list Department, University, City, Region, Country
% Industry affiliations should list Company, City, Region, Country

% you can specify symbols, otherwise they are numbered in order
% ideally, you should not use this facility. affiliations will be numbered
% in order of appearance and this is the preferred way.
% \icmlsetsymbol{equal}{*}

\begin{icmlauthorlist}
\icmlauthor{Martin Raison}{fb}
\icmlauthor{Pierre-Emmanuel Mazar\'e}{fb}
\icmlauthor{Rajarshi Das}{umass}
\icmlauthor{Antoine Bordes}{fb}
\end{icmlauthorlist}
\icmlaffiliation{fb}{Facebook AI Research, Paris, France}
\icmlaffiliation{umass}{University of Massachusetts, Amherst, USA}
\icmlcorrespondingauthor{Martin Raison}{raison@fb.com}
%\icmlcorrespondingauthor{Pierre-Emmanuel Mazar\'e}{pem@fb.com}

% You may provide any keywords that you 
% find helpful for describing your paper; these are used to populate 
% the "keywords" metadata in the PDF but will not be shown in the document
% \icmlkeywords{boring formatting information, machine learning, ICML}

\vskip 0.3in
]

% this must go after the closing bracket ] following \twocolumn[ ...

% This command actually creates the footnote in the first column
% listing the affiliations and the copyright notice.
% The command takes one argument, which is text to display at the start of the footnote.
% The \icmlEqualContribution command is standard text for equal contribution.
% Remove it (just {}) if you do not need this facility.

\printAffiliationsAndNotice{}  % leave blank if no need to mention equal contribution
%\printAffiliationsAndNotice{\icmlEqualContribution} % otherwise use the standard text.

\begin{abstract} 
%The purpose of this document is to provide both the basic paper template and
%submission guidelines. Abstracts should be a single paragraph, between 4--6 sentences long, ideally.  Gross violations will trigger corrections at the camera-ready phase.

This paper aims at improving how machines can answer questions directly from text, with the focus of having models that can answer correctly multiple types of questions and from various types of texts, documents or even from large collections of them.
To that end, we introduce the \us model that uses a new way to relate a question to a textual context by weaving layers of recurrent networks, with the goal of making as few assumptions as possible as to how the information from both question and context should be combined to form the answer. 
We show empirically on six datasets that \us performs well in multiple conditions. For instance, it produces solid results on the very popular SQuAD dataset \citep{rajpurkar2016squad}, solves almost all bAbI tasks \citep{weston2015towards} and greatly outperforms state-of-the-art methods for open domain question answering from text \citep{chen2017reading}.

\end{abstract}

\section{Introduction}

Being able to answer any question from large collections of unstructured text would give machines the ability to use as knowledge sources the huge amounts of varied information available in online encyclopedias (Wikipedia, etc.), but also in news articles, forums, blogs or social media posts.
In theory, this unrestricted access to rich and dynamic information should lead to improved answering abilities. This comes, however, at the expense of having to solve a much harder problem than when using knowledge stored in structured databases. Here systems cannot rely on any ontology, and instead have to search large collections of documents and carefully read them to find the answers.

\citet{chen2017reading} proposed the \drqa system to tackle this problem through machine reading at scale, that is, answering questions using spans of tokens extracted from Wikipedia. The \drqa pipeline system is composed of two modules: a document retriever and a document reader. The retriever is an information retrieval system based on TF-IDF matching that returns a subset of documents given a question. The reader is a model for the task of answering questions given a textual context, or \emph{machine reading}, that uses bi-directional LSTMs (BiLSTMs) to get candidate answers from paragraphs and aggregate them afterwards. 

Both modules perform quite well separately, but their combination experiences several limitations as illustrated by the following performance on the development set of the SQuAD dataset \citep{rajpurkar2016squad}. When provided with the paragraph containing the answer, the reader can respond correctly to around 70\% of the questions. When provided with the \emph{entire} Wikipedia article containing the answer, however, this performance drops to 50\%. This performance hit is compounded when integrated with the imperfect retriever. Considering the top 5 retrieved articles causes the performance of the reader to drop further to a final accuracy below 30\% for \drqa. 
In this work, we focus on improving the reader to make it more general, more robust to longer contexts and hence more accurate overall.

Many recent advances in machine reading are done on the SQuAD dataset, with impressive results: the top performing methods on the leaderboard can compete with results of human subjects. However, the SQuAD dataset has its specificities, limitations and does not cover the whole range of question answering; there is a risk to overspecialize models for it. 
Recurrent neural networks and attention mechanisms are key components of those models. BiLSTMs are used for encoding questions and contexts into continuous representations, while attention is used for connecting them by building question-aware context representations and context-aware question representations. Self-attention is also used to give the model an opportunity to match different parts of the context with each other. Building such architectures requires multiple choices to decide how the various types of attention and recurrent layers should be combined together. 
For instance, the top published models on SQuAD\footnote{as of February 2018.} \citep{san,fusionnet,dcn+} all use at least four types of attention mechanisms in their models, whether through attention, co-attention or self-attention but have only been tested on SQuAD or variants \citep{jia2017adversarial}.

With the objective of being flexible to various types of questions, contexts and tasks, this paper introduces \us, a machine reading model that does not require any attention mechanism for building the representations for questions and contexts. Instead, \us relies on multiple layers of BiLSTMs that are woven to co-encode both questions and contexts simultaneously and hence learn to make both representations deeply interconnected, without having to specify how in the architecture.
These representations are then used by an answering layer inspired by Memory Networks \citep{sukhbaatar2015end} to produce the final response.
We test \us on 6 different datasets for question answering including SQuAD and show that it performs remarkably well in various conditions.

In addition to strong results on SQuAD, \us is able to solve 17/18 bAbI tasks \citep{weston2015towards} that test different answering skills. It can even solve tasks that require to output words that do not belong to the context, something that limits most machine reading models. 
\us is also robust to finding responses in collections of short contexts as illustrated by its state-of-the-art results obtained on the QAngaroo WikiHop dataset \citep{welbl2017qangaroo}.
As a result, its versatility and robustness make \us a very promising reader component for machine reading at scale, better suited than the original \drqa reader was. Indeed, while \us is around 5\% better than \drqa for answering from paragraphs, when we use it in the pipeline for answering from the full Wikipedia, it boosts the overall performance by more than 15\%.

\section{Related Work}

The machine reading task of learning to automatically answer questions given a provided piece of text (news article, fictional story, Wikipedia snippet, etc.) has been making great progress in recent years thanks to the creation of datasets like MCTest \citep{richardson2013mctest}, QACNN/DailyMail \citep{nips2015hermann}, CBT \citep{hill2015goldilocks}, WikiQA \cite{yang2015wikiqa}, bAbI \citep{weston2015towards}, SQuAD \citep{rajpurkar2016squad} or QAngaroo \citep{welbl2017qangaroo} --- as well as initiatives to group them together like ParlAI \citep{miller2017parlai}.

Most recent advances in machine reading expect that a supporting document is provided when a question is asked; they are not suited for the open-domain scenario in which one has to search through large databases to answer. 
Until recently, open-domain question answering had been mostly addressed through the task of answering from structured knowledge bases such as WikiData \citep{vrandevcic2014wikidata} or DBpedia \citep{auer2007dbpedia}. However, the limitations of knowledge bases (missing information, rigid schema, imperfect information, etc.) and the recent advances in machine reading have allowed new progress in the original trend of work of answering from large collections of unstructured text.
Hence, new resources mixing questions with textual pieces of evidence returned by a search engine have been recently proposed. These include MSMARCO \citep{nguyen2016ms} where questions sampled from real
anonymized user queries are paired with real web documents retrieved by the Bing search engine; SearchQA \citep{dunn2017searchqa} that mixes question-answer pairs from the J! Archive with text snippets retrieved by Google; and TriviQA \citep{JoshiTriviaQA2017} that includes question-answer pairs authored by trivia enthusiasts along with independently gathered evidence documents. In this paper, we follow the setting used in \citep{chen2017reading,wang2017r} to be comparable with the results therein and hence use open-domain versions of SQuAD, WebQuestions \citep{berant2013semantic}, WikiMovies \citep{miller2016key} and CuratedTREC \citep{baudivs2015modeling}. Our method could be adapted to MSMARCO, TriviaQA or SearchQA, though this is left as future work.

Numerous neural models have been proposed to jointly encode questions and textual evidence for machine reading. Most of them follow the same general structure of first using recurrent architectures to encode questions and contexts separately and then using multiple types of attention mechanisms to connect them before running a final answering step. In this paper, we instead propose to co-encode questions and contexts with the same recurrent layers directly and use an answering step inspired by the multi-hops approach used in \citep{sukhbaatar2015end,san}.

Most of the recent architectures have been developed and tested on a single or very few datasets, usually bAbI, QACNN and recently SQuAD, which leaves some questions regarding their capacity to adapt to multiple types of conditions. For instance, \citet{nips2015hermann} and \citet{kadlec2016text} tested their models in the Cloze setting of QACNN/DailyMaill and CBT only. \citet{sukhbaatar2015end} and \citet{seo2016query} focused on solving the bAbI tasks. Recently, a large body of work have been proposed to tackle the SQuAD dataset only e.g. in \citep{dhingra2016gated,wang2017gated,dcn+,conductornet,reinforced-mnemomic,fusionnet,san}.
Besides, almost none of them has been applied to a setting requiring to answer from a large collections of documents. Some of the exceptions are the Reinforced Mnemonic Reader \citep{reinforced-mnemomic} that has been tested on SQuAD and TriviaQA or
\drqa \citep{chen2017reading} and the Reinforced Reader-Ranker \citep{wang2017r} that both attempted to tackle the full pipeline problem combining retrieving relevant documents from Wikipedia and reading them to answer.

We show in our experiments in Section~\ref{sec:exp} that \us can perform well in various conditions from the bAbI tasks  to SQuAD while also being applied in an open-domain setting where it outperforms by a vast margin both \drqa and the Reinforced Reader-Ranker. The latter method focuses on improving the connection between the retrieving and the reading modules through reinforcement learning but our results demonstrate major gains can already be obtained by improving the reading module alone.

\begin{figure*}[t]
\begin{center}
\includegraphics[width=0.9\textwidth]{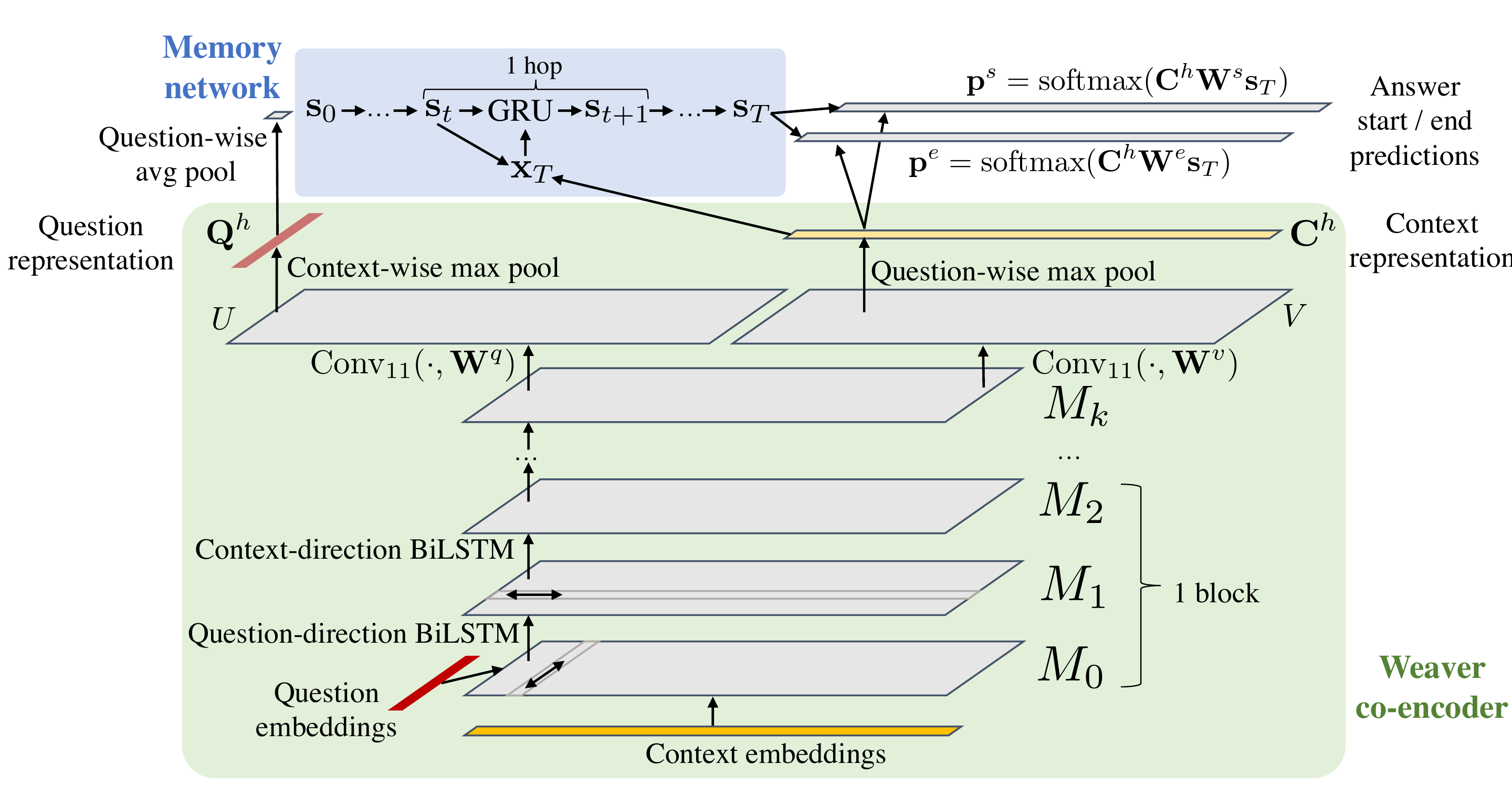}
\end{center}
\caption{\label{fig:arch} \us architecture. At the bottom of the figure, the \us co-encoder encodes the question and context into variable-length representations through a series of alternating BiLSTM layers. We reduce the question representation into a fixed size and use it as the initial state of our memory network (top left). At each hop, the memory network will update its state $\mathbf{s}_t$ based on the context representation $\mathbf{C}^h$. We then compute a bilinear projection of the final state of the memory network and the context representation to predict independently the start and end position of the answer span (top right).}
\end{figure*}

\section{The \us Architecture} 

\us is a model for machine reading that attempts to answer a question $q$ by identifying a response span in a textual context $c$.
The architecture of \us is composed of four parts: (i) embedding of the input words, (ii) deep co-encoding of $q$ and $c$, (iii) a memory network step and (iv) a final answer prediction. An overall architecture diagram can be found in Fig.~\ref{fig:arch}.

\subsection{Input Embeddings}

We first tokenize the question and context into individual words, and associate each word with a word embedding. The question and context are thus represented as sequences of respectively $m$ and $n$ word embeddings $\mathbf{q}_1, \mathbf{q}_2, ..., \mathbf{q}_m$ and $\mathbf{c}_1, \mathbf{c}_2, ..., \mathbf{c}_n$. Unless otherwise noted, we use 300-dimensional FastText word embeddings trained on Common Crawl \citep{mikolov2017advances} and keep them fixed during training. Out-of-vocabulary words are represented with a fixed randomly initialized vector.

As in \citep{chen2017reading}, we also associate each context word with an exact match feature and token features (part-of-speech, named entity recognition tags, term frequency). We denote these additional features $\mathbf{c}^{\text{extra}}_j$.

\subsection{Question and Document Co-encoding}

We consider a 2D map of size $m\times n$. For each coordinate $(i, j)$ in the map, we combine the question and context feature vectors with a function
\begin{equation}
f: (\mathbf{q}_i, \mathbf{c}_j, \mathbf{c}_j^{extra}) \mapsto [\mathbf{q}_i;\mathbf{q}_i-\mathbf{c}_j;\mathbf{q}_i^\intercal \mathbf{c}_j;\mathbf{c}^{\text{extra}}_j]
\label{eqn:input}
\end{equation}
where $[\cdot;\cdot]$ represents vector concatenation. This yields a 3D tensor $M_0$ of size $m\times n\times d$, where $d$ is the dimensionality of the concatenated vector. 

A simpler definition of $f$ would be to just concatenate $\mathbf{q}_i$, $\mathbf{c}_j$ and $\mathbf{c}_j^{\text{extra}}$, but given that $\mathbf{q}_i$ and $\mathbf{c}_j$ live in the same vector space, using $\mathbf{q}_i-\mathbf{c}_j$ and $\mathbf{q}_i^\intercal \mathbf{c}_j$ explicitly makes it easier for the model to match tokens together. Note that $f$ does not require any trainable parameters.

We then transform $M_0$ into a higher-level representation using recurrent layers. Since recurrent layers are typically one-dimensional, we propose to apply them alternatively along the question direction and the context direction. We call the sequence of one question-wise operation and one context-wise operation a \emph{block}, and we propose to stack several blocks on top of each other.

More precisely, for the first block:
\begin{enumerate}
\item
We first slice the input tensor $M_0$ in the context direction, giving us $n$ slices of size $m\times d$. Considering each slice as a sequence of length $m$ and input size $d$, we apply a single BiLSTM to each of those $n$ slices. For each slice, the BiLSTM yields an output of size $m\times 2h$ (where $h$ is the hidden layer size). We concatenate all those outputs back together to get a 3D tensor $M_1$ of size $m\times n\times 2h$.
\item
Similarly, we then slice $M_1$ in the question dimension to get $m$ slices of size $n\times 2h$. Considering each slice as a sequence of length $n$ and input size $2h$, we apply another BiLSTM to each of those $m$ slices. For each slice, the BiLSTM output is of size $n\times 2h$. We concatenate all those outputs back together to get a 3D tensor $M_2$ of size $m\times n\times 2h$.
\end{enumerate}

We can stack such blocks by repeating these steps several times replacing $d$ with $2h$ in the first step. The final output $M_k$ has size $m\times n\times 2h$. All BiLSTMs used in this case have different sets of weights.
We also add residual connections to all consecutive layers except the first one (which has different dimensionality), i.e. we add $M_i$ to $M_{i+1}$ before feeding to the next layer.
Given the shape of the $M_0$ tensor, an approach using 2D convolutions at each layer would seem more intuitive, however we found the alternation of recurrent layers proposed here to perform significantly better on the development set of SQuAD.

From the $M_k$ tensor we obtain the final representations for the question and the context in the following way.
We compute two 3D tensors $U = \textup{Conv}_{11}(M_k, \mathbf{W}^q)$ and  $V = \textup{Conv}_{11}(M_k, \mathbf{W}^v)$ as the result of convolutions of size 1 over $M_k$, where $\mathbf{W}^q$ and $\mathbf{W}^v$ are learned matrices of size $2h \times 2h$. 
These two different convolutions are important to distinguish questions and context features.
We then derive a fixed-size representation for each token in the context by applying to $V$ a max-pooling along the question dimension. This produces the \emph{context representation}, a matrix $\mathbf{C}^h$ of size $n\times 2h$. Similarly, we compute a fixed-size representation of each token in the question by applying max-pooling along the context dimension to $U$, yielding the \emph{question representation} a matrix $\mathbf{Q}^h$ of size $m \times 2h$.

The two representations $\mathbf{Q}^h$ and $\mathbf{C}^h$ are of variable length and maintain a sense of locality within the original context or question. For machine reading, we use the locality of information on the context representation to predict the likelihood for each token in the context to be the start or the end of the answer to the question.

In practice, recurrent layers are more efficiently applied to batches of sequences instead of isolated ones. By transposing and reshaping $M_i$ appropriately at each layer, we can apply the BiLSTMs within blocks to all slices in parallel. This makes the model straightforward to implement.

\subsection{Memory Network}

We can directly use $\mathbf{Q}^h$ and $\mathbf{C}^h$ to predict the probability of each token to be the start or the end of an answer. However, multiple previous works have shown that it could be beneficial for the question to attend iteratively to the context before predicting an answer. The mutual conditioning of context and question representations is not sufficient to represent such an iterative process. We therefore add a multi-hop attention process before answering.

We apply an end-to-end memory network as introduced by \citet{sukhbaatar2015end}. At each hop $t$, the memory network computes its updated state $\mathbf{s}_{t+1}$ depending on the context representation $\mathbf{C}^h$ and its previous state $\mathbf{s}_t$. We first apply an average-pooling to the question representation $\mathbf{Q}^h$ along the question dimension in order to reduce it to a single vector $\mathbf{s}_0$ of length $2h$ that is used as the initial state of the memory network.
As proposed in \citep{san}, each hop updates the state through a Gated Recurrent Unit (GRU) \cite{cho2014gru}. We thus apply the following update rule for each hop $t$:
\begin{align*}
\mathbf{x}_t &= \mathbf{C}^h\mathbf{W}^c\textup{softmax}(\mathbf{C}^h\mathbf{W}^h\mathbf{s}_t) \\
\mathbf{s}_{t+1} &= \textup{GRU}(\mathbf{x}_t, \mathbf{s}_t)
\end{align*}

with $\mathbf{W}^h$ and $\mathbf{W}^c$ learned matrices of size $2h \times 2h$.

The total number of hops $T$ is an hyperparameter typically set between 1 and 5 in our experiments. Having 0 hop corresponds to bypassing the memory network entirely.

\subsection{Answer Prediction}

Unlike \citet{san} that average all states $\mathbf{s}_t$ of the GRU to compute the probability of each answer span in the context, we only use the final state $\mathbf{s}_T$ to do so because we found it to be more accurate in practice. We define:
\begin{align*}
\mathbf{p}^s &= \textup{softmax}(\mathbf{C}^h\mathbf{W}^s\mathbf{s}_T) \\
\mathbf{p}^e &= \textup{softmax}(\mathbf{C}^h\mathbf{W}^e\mathbf{s}_T)
\end{align*}
where $\mathbf{W}^s$ and $\mathbf{W}^e$ are learned matrices of size $2h \times 2h$, and $\mathbf{p}^s_i$ (resp. $\mathbf{p}^e_i$) represents the probability that position $i$ is the start (resp. the end) of the answer span with $0\leq i< n$.

At inference, we use unnormalized exponential instead of softmax to make scores compatible across paragraphs in one or several documents. We select the span $[i,j]$ which maximizes $\mathbf{p}^s_i\mathbf{p}^e_j$ for $i \leq j \leq i + 15$.

\section{Experimental Setup}

\subsection{Datasets}

We test our model's ability to answer questions with various types of context, synthetic stories, paragraphs, documents, and full encyclopedia (Wikipedia), using the datasets described in this section. Two evaluation metrics are used depending on the dataset: the exact string match (EM) between the predicted span and the answer string and the F1 score, which measures the harmonic mean of precision and recall at the token level. All scores reported in this paper are percentages. In the full Wikipedia setting, we use the open-domain question answering datasets and the same Wikipedia dump\footnote{We use the english-language Wikipedia dump of 2016-12-21 (\url{https://dumps.wikimedia.org/enwiki/latest}), which contains 5,075,182 articles consisting of 9,008,962 unique uncased token types.} as \citep{chen2017reading,wang2017r}.
Statistics of the datasets are given in Table~\ref{tab:data-stats}.

\begin{table}[t]
\begin{center}
\begin{tabular}{l|c@{\,\,}c@{\,\,}c}
Dataset & \multicolumn{2}{c}{Train} & Test  \\
& Plain & DS &  \\
\hline
bAbI &  10,000$^{\ddagger}$ & - & 1,000$^{\ddagger}$ \\
Wikihop &  43,738 & - & 5,129$^{\dagger}$ \\
SQuAD &  87,599 & - & 10,570$^{\dagger}$ \\
CuratedTREC &  1,486$^{*}$ & 2,643 & 694 \\
WebQuestions &  3,778$^{*}$ & 6,308 & 2,032 \\
WikiMovies &  96,185$^{*}$ & 100,528 & 9,952 \\
\end{tabular}
\end{center}
\caption{\label{tab:data-stats} Number of questions for each dataset used in this paper. DS: distantly supervised training data.
$^{*}$: These training sets are not used as is because no paragraph is associated with each question.
$^{\dagger}$: Corresponds to Wikihop and SQuAD development set. $^{\ddagger}$:~All figures for bAbI tasks refer to a single task.}
\end{table}

\paragraph{BAbI Tasks} This dataset consists in 20 simple dialog tasks. In line with previous literature on the dataset, we consider a task solved if the EM accuracy reaches $95$ on the validation set. 
We select the set of tasks for which the answer is a single word, hence excluding tasks 8 and 19.

\paragraph{QAngaroo Wikihop} This dataset was introduced in \cite{welbl2017qangaroo}. In this dataset, questions are not sentences but consist in the concatenation of a subject and a knowledge base relation, e.g. \emph{place\_of\_birth caspar john} would mean \emph{Where was Caspar John born?}. We do not use any special approach for such inputs apart from a forced tokenization around the \_ character. In particular, the model does not learn representations for relations but only for the individual words that appear in the relation's text form, e.g. \emph{place of birth}. Documents are sequences of \emph{supports} that contain information coming from several Wikipedia pages. We concatenate all the supports for a given example and use it as a single context.

\paragraph{SQuAD} The Stanford Question Answering Dataset \citep{rajpurkar2016squad} contains 87k questions over Wikipedia for training and 10k for development, with a large hidden test set which can only be accessed by the SQuAD creators.
Each example is composed of a paragraph extracted from a Wikipedia article and an associated human-generated question. The answer is always a span from this paragraph and a model is given credit if its predicted answer matches it.

\paragraph{CuratedTREC} This dataset is based on the benchmarks from the TREC question answering tasks that have been curated by \citet{baudivs2015modeling}. We use the large version, which contains a total of 2,180 questions.

\paragraph{WebQuestions} \citet{berant2013semantic} built this dataset around the task of answering questions from the Freebase knowledge base. It was created by crawling questions through the Google Suggest API, and then obtaining answers using Amazon Mechanical Turk.
Each answer has been converted by \citet{chen2017reading} to text by using entity names so that the dataset does not reference Freebase IDs and is purely made of plain text question-answer pairs.

\paragraph{WikiMovies} \citet{miller2016key} collected 96k question-answer pairs in the domain of movies. Originally created from the OMDb and MovieLens databases, the examples are built such that they can also be answered by using a subset of Wikipedia as the knowledge source (the title and the first section of articles from the movie domain).

\subsection{Distant Supervision}

Unlike recent machine reading comprehension datasets, CuratedTREC, WebQuestions and WikiMovies  only contain question, answer pairs and lack any supporting documents. To gather supporting documents we resort to distant supervision \cite{mintz2009distant}. We follow the same distant supervision setup as \cite{chen2017reading} but since we are interested in machine reading over long documents, we associate full documents instead of single paragraphs, and thus increase the character limit to 100,000 instead of 1,500 originally. We use those distantly supervised training sets to fine-tune a model trained on SQuAD. Corresponding results are given in Table~\ref{tab:fullp}.

\subsection{Implementation Details}

For training, we batch together examples with similar document sizes, padding all matrices in the question and context directions when necessary. Within each epoch, mini-batches are shuffled randomly. \us is trained via optimizing the sum of the cross-entropy losses for the start and end points of the answer span for each training example. To this end, we use Adamax with a learning rate of $0.002$ on the padded mini-batches. Weights are initialized randomly according to a Gaussian of mean 0 and variance $1/n$ where $n$ is the number of the neuron's input channels. We apply dropout at a rate of $0.2$ to the output of all LSTMs as well as dropout at a rate of $0.3$ on $M_0$.
Model selection is done using the validation set of each dataset.

On all datasets but bAbI, we use the Stanford CoreNLP toolkit \citep{manning2014stanford} to tokenize the input documents and generate part-of-speech and named entity recognition features. \us is implemented in PyTorch.\footnote{\url{http://pytorch.org}.}

For datasets providing a list of candidate answers (QAngaroo WikiHop, WebQuestions and WikiMovies), we restrict the answer span to be in this list during prediction.

\section{Experimental Results} \label{sec:exp}

This section presents the performance of \us on a wide range of tasks, from reasoning on short synthetic stories (bAbI) to open domain question answering on Wikipedia.

\begin{table}[t]
\begin{center}
\begin{tabular}{r|cccc}
Task & \multicolumn{2}{|c}{\drqa} & \multicolumn{2}{c}{\us} \\
& single & multi-task & single & multi-task \\
\hline
1 & 100 & 100 & 100 & 100 \\
2 & 98.1 & \rd{46.6} & 99.2 & 99.7 \\
3 & \rd{45.4} & \rd{55.6} & 99.3 & 99.7 \\
4 & 100 & 96.3 & 100  & 100 \\
5 & 98.9 & 98.1 & 99.8 & 99.8 \\
6 & 98.4 & 99.9 & 100& 100 \\
7 & 100 & 99.1  & 99.8& 100 \\
9 & 100 & 99.8 & 100 & 100 \\
10 & 99.7 & 100 & 100 & 100 \\
11 & 100 & 100 & 100 & 100 \\
12 & 100 & 100 & 100 & 100 \\
13 & 100 & 100 & 100 & 100 \\
14 & 99.8 & 96.0  & 99.9& 99.9 \\
15 & 100 & \rd{57.6}  & 99.5& 100 \\
16 & \rd{47.7} & \rd{44.5} & \rd{53.3} & \rd{49.0} \\
17 & \rd{94.9} & \rd{60.4} & 100& 100 \\
18 & 100 & \rd{93.4}  & 100& 100 \\
20 & 100 & 100  & 100& 100 \\
\hline
Failed & 3 & 6 & 1 & 1 \\
\end{tabular}
\caption{\label{tab:babi} Test accuracies of \us and DrQA on bAbI-10k. We did not test on tasks 8 and 19 because they require to answer with multiple words but \drqa and \us are not designed to do so.}
\end{center}
\end{table}

\subsection{Reasoning on Synthetic Stories}

On this dataset, we do not use fixed pretrained embeddings but learn embeddings of dimension 128 based on a random initialization. With a 1-block co-encoder with $h=128$ and 3 hops, \us solves 17 out of the 18 tasks that are relevant for it, as shown in Table~\ref{tab:babi}. 
It can solve them both in the single setting (each task trained separately) and the multi-task setting (all tasks trained together), while DrQA could solve only 15 single tasks and 12 in a multi-task setting.
Note that solving all bAbI tasks is not trivial, since only two dedicated models did it \citep{seo2016query,henaff2016tracking} to the best of our knowledge.

Several bAbI tasks require to answer with words that do not belong to the context such as ``yes'', ``no'' or ``maybe''. Standard span-based machine reading models cannot readily do that. We overcome this limitation here by prepending the context with a list of extra words that can be picked by the model to answer. This list is made as the union of all answers to these tasks on the training set, which forms a sequence of 18 words present at the beginning of every context.
As shown by the good results in the multi-task setting for which we prepend those extra words for all examples, \us can learn to use them only when it is relevant.

The number of hops has a large impact on the performance on some tasks such as Task 3. As shown in Table~\ref{tab:hops}, this task could not be solved without one memory network hop, and the performance increases steadily with the number of hops. 
This effect can also be observed on other datasets even though it is most visible on the bAbI dataset thanks to its synthetic nature.

\subsection{Answering from Paragraphs}

\subsubsection{QAngaroo Wikihop}

We use a 1-block co-encoder with $h = 200$ and between 1 and 5 memory network hops. 
As shown in Table~\ref{tab:hops}, \us can achieve up to 64.1 EM accuracy on the development set, with a single hop being sufficient to achieve the best performance. \us achieves 65.3 EM accuracy on the hidden test set, which beats the previous state of the art by a 5-point margin.

\subsubsection{SQuAD}

We use a 2-block co-encoder with $h = 200$ and train for 20 epochs. Table~\ref{tab:squad} shows that our model achieves 82.8 F1 on the hidden test set, which is comparable to the best published architectures: it is better than Conductor-net \citep{conductornet} and Reinforced Mnemonic Reader (M-Reader + RL) \citep{reinforced-mnemomic} and outperformed by DCN+ \citep{dcn+}, FusionNet \citep{fusionnet} and SAN \citep{san}. 
It is worth noting that all those methods have only been tested on SQuAD, except M-Reader + RL that was also applied on TriviaQA.

The addition of the memory network (with 1 or more hops) improves the F1 performance by 0.9 points compared to using directly the question and context representations. We empirically find that a deeper memory network (i.e. more hops) requires a lighter co-encoding (i.e. fewer blocks) to achieve the same performance. We also find that the variability due to the random network initialization implies a standard deviation of the results of 0.2 F1 points.

We benchmarked several architecture variants. We added a character-level LSTM to embed out-of-dictionary words and incorporate this data into the $M_0$ layer. This addition did not improve performance measurably. We can attribute this to the fact that most words in SQuAD are covered by the FastText word embeddings: in fact, out-of-dictionary words represent only a total of 61k words or 1.5\% of the full corpus (which comprises 3.9M words, duplicates included). Most of these missing words may  be interpretable thanks either to surrounding words or to part-of-speech and named entity recognition features.
In other experiments, we also added self-attention layers but they did not bring any improvement to what the plain co-encoder of \us can do.

\begin{table}
\begin{center}
\begin{tabular}{l|cccc}
& \multicolumn{2}{c}{Dev set} & \multicolumn{2}{c}{Test set} \\
 & EM & F1 & EM & F1 \\
\hline
DrQA & 69.5 & 78.8 & 70.7	& 79.3 \\
\hline
Conductor-net & 72.1 & 81.4 & 72.6 & 81.4 \\
M-Reader+RL & 72.1 & 81.6 & 73.2 & 81.8 \\
DCN+ & 74.5 & 83.1 & 75.1 & 83.1 \\
FusionNet & 75.3 & 83.6 & 76.0 & 83.9 \\
SAN & \bf 76.2 & \bf 84.1 & \bf 76.8 & \bf 84.4 \\
\hline
\us & 74.1 & 82.4 & 74.4 & 82.8 \\
\end{tabular}
\caption{\label{tab:squad} Results of single models on SQuAD. We include as baselines the best published models as of February 2018.}
\end{center}
\end{table}

\begin{table}
\begin{center}
\begin{tabular}{c|ccc}
 Nb of & bAbI Task 3 & WikiHop & SQuAD\\
 Hops  & (EM) & (EM) & (Dev F1)\\
\hline
0 & 66.5 & 61.2 & 81.9 \\
1 & \bf 99.0 & \bf 64.1 & 81.9 \\
3 & \bf 99.3 & 63.4 & 82.0 \\
5 & \bf 99.5 & 63.0 & \textbf{82.4} \\
7 & - & - & 82.1 \\
9 & - & - & 81.8
\end{tabular}
\caption{Impact of the number of hops in the memory network final step of \us (EM: exact match accuracy).\label{tab:hops}}
\end{center}
\end{table}

\subsection{Answering from Documents}

We then test our model in the more difficult setting where a full document is given as context to answer the question. On SQuAD, documents are on average 40 times larger than individual paragraphs. All paragraphs of the document are processed independently and we pick the answer span with the maximum score across all paragraphs.

We compare with \drqa trained on paragraphs and also with an improved version, which we term $\text{\drqa}^\star$\label{drqastar}, that has the following changes:
\begin{itemize}
\item
We use the same FastText pretrained embeddings as \us. \citep{mikolov2017advances} showed that this improves the performance of \drqa on SQuAD, compared to using the GloVe embeddings \citep{glove} as in the original \drqa model.
\item
The relatively small size of the \drqa model makes it practical to read the whole document contiguously, as opposed to processing paragraphs independently. This also helps it achieve better results.
\end{itemize}

For both models, training on full documents is critical to get good performance on this task. To speed up training of the \us model, instead of processing all paragraphs, we pick the paragraph containing the correct answer and sample 5 other paragraphs at random from the document. We take the final softmax across all paragraphs. As seen on Table~\ref{tab:docs}, our model gains 6 points in F1 score compared to the model only trained to answer on single paragraphs and also significantly outperforms the baselines.

\begin{table}
\begin{center}
\begin{tabular}{l|ll|cc}
  & Train & Test & EM &  F1 \\
\cline{1-5}
\drqa & paragraph & full doc. & 49.4 & 58.0 \\
$\text{\drqa}^\star$ & paragraph & full doc. & 59.1 & 67.0 \\
$\text{\drqa}^\star$ & full doc. & full doc. & 64.7 & 73.2 \\
\cline{1-5}
\us & paragraph & full doc. & 60.6 & 69.7 \\
\us & full doc. & full doc. & \bf 67.0 & \bf 75.9 \\
\end{tabular}
\caption{\label{tab:docs} Document-scale results with training and testing on SQuAD (using the dev set for evaluation). For each architecture, we indicate the setting used for training and testing (single paragraph or full document). The $\text{\drqa}^\star$ model refers to an improved version of \drqa (see section \ref{drqastar} for details).}
\end{center}
\end{table}

\subsection{Answering from the Full English Wikipedia}

\begin{table*}[t]
\begin{center}
\begin{tabular}{ll|cccc}
\multicolumn{1}{c}{} &  & SQuAD & CuratedTREC & WebQuestions & WikiMovies \\
\hline
YodaQA   & {\it - addtl sources} & - & 31.3 & \bf 39.8 &-  \\ 
\drqa  & {\it - SQuAD train} & 27.1 & 19.7 & 11.8 & 24.5  \\
 & {\it - fine-tuning} & 28.4 & 25.7 & 19.5 & 34.3   \\
$\text{\drqa}^\star$ & {\it - SQuAD train} & 39.5 & 21.3 & 14.2 & 31.9  \\
 & {\it - fine-tuning} & 40.4 & 28.8 & 24.3 & \bf 46.0   \\

\rrr  & {\it - fine-tuning} & 29.1 & 28.4 & 17.1 & 38.8 \\ 
\hline

\us & {\it - SQuAD train} & \bf 42.3 & 21.3 & 13.0 & 33.6 \\
 & {\it - fine-tuning} & - & \bf 37.9 & 23.7 & 43.9   \\
\end{tabular}
\caption{\label{tab:fullp} Results when answering questions using the full English Wikipedia (top-1 EM accuracy, using SQuAD evaluation script)}
\end{center}
\end{table*}

To answer questions from all of Wikipedia, we use a 2-step pipeline similar to \citep{chen2017reading}:
\begin{enumerate}
\item
We first use the same retriever as \citep{chen2017reading} to find $k$ documents related to the question.\footnote{\url{https://github.com/facebookresearch/DrQA}} This retriever uses bigram hashing and TF-IDF matching to score documents relative to the question.
\item
We then run the \us model (trained on full documents) as a document reader on the top $k$ retrieved documents, and return the span with the maximum score across all documents.
\end{enumerate}

\begin{figure}
  	\centering
   	\resizebox {\columnwidth} {!} {
		\begin{tikzpicture}
  			\begin{axis}[
  				legend pos=south east,
  				xlabel=Number of documents retrieved ($k$),
				ylabel=EM score,
                xmin=0,
                ymin=0,
    			xtick= {1, 5, 10, 15, 25}]
    			\addplot [teal,mark=*]  table[x=k,y=drqapp] {\fullpipelinedata};
	    		\addlegendentry{\us}
    			\addplot [orange,mark=x] table[x=k,y=drqa] {\fullpipelinedata};
	    		\addlegendentry{\drqa}
  			\end{axis}
		\end{tikzpicture}
	}
	\caption{\label{fig:fullpipeline} Impact of the number of documents retrieved for each question in the full Wikipedia setting.}
\end{figure}
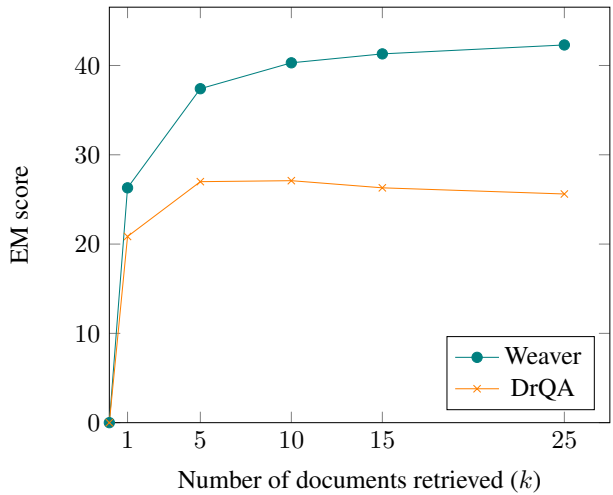

All results are reported in Table~\ref{tab:fullp}, where we compare with \drqa, $\text{\drqa}^\star$, the Reinforced Reader-Ranker \citep{wang2017r} and YodaQA \citep{baudivs2015yodaqa}.
In addition to showing better performance than the original \drqa model on a single full document, \us is also able to handle more documents properly. While the EM score for DrQA reaches its maximum with 10 documents, we are able to increase the performance of the full system with \us as a reader by increasing the number of retrieved documents up to 25, see Fig.~\ref{fig:fullpipeline}. This allows to reach a final EM accuracy of 42.3 on SQuAD, which is more than 12 points better than the previous best reported performance.

We also evaluate our model on CuratedTREC, WebQuestions and WikiMovies, first by directly testing the model trained on the SQuAD training set, and then after fine-tuning this model on each dataset independently.
For the fine-tuning stage on a given dataset, we take our best model on the SQuAD full document setting and train it for 1 epoch on the distantly supervised corresponding training set (20 epochs for WebQuestions).
Performance on CuratedTREC presents a significant improvement over the state of the art (6.6 EM gain). Note that we only count an exact match if the provided regular expression matches the full span returned by the model. If we also count an exact match when the regex matches a substring (as intended by the dataset authors\footnote{\url{https://github.com/brmson/dataset-factoid-curated}}), we obtain 39.2 EM with the model trained on SQuAD and 43.8 EM with the fine-tuned model. Performance on WikiMovies also increases significantly although our improved baseline $\text{\drqa}^\star$ performs better when fine-tuned on that dataset.

Performance on WebQuestions remains lower than that of YodaQA but, in addition to Wikipedia, YodaQA uses Freebase as additional knowledge source, which gives a key advantage on this dataset created from Freebase.

\pgfplotstableread{
k	drqa	drqapp
0   0       0
1	20.85	26.30
5	26.99	37.4
10	27.1	40.3
15	26.30	41.3
25	25.61	42.3
}\fullpipelinedata

\subsection{Ablation Study}

\begin{table}
\begin{center}
\begin{tabular}{c|cc}
 Model  & Dev EM & Dev F1\\
\hline
\bf \us & \bf 74.1 & \bf 82.4 \\
CatQC & 71.8 & 80.8 \\
CatQCDotProduct & 72.3 & 81.3 \\
NoRnn & 22.7 & 33.0 \\
NoConv11 & 71.8 & 81.1 \\
NoMemNet & 72.9 & 81.9
\end{tabular}
\caption{Ablation study on SQuAD dev set. Ref: reference model. CatQC: we substitute the input representation mentioned in Eq. \eqref{eqn:input} by a direct concatenation of question and context embeddings. CatQCDotProduct: starting from the CatQC architecture, we add back the dot-product between question and context embedding to the input representation. NoRnn: weaved RNN layers are replaced with a linear projection. NoConv11: we remove the convolution layer. NoMemNet: we set the number of hops to 0 in the memory network. \label{tab:abs}}
\end{center}
\end{table}

To verify that the prediction performance improvements come from the co-encoding module, we conduct an ablation study on the SQuAD dev set. We report results in Table \ref{tab:abs}. Architecture choices such as the exact input encoding or the presence of modules such as the convolutional layer or the memory network play a minor role in the prediction performance. However, a model without the stacked recurrent neural networks reaches only 33 F1. This shows that the RNNs applied in a weaving pattern are the major factor that enables a good prediction performance.

\section{Conclusions and Future Work}

In this paper, we introduced \us, a novel way of building question and context representations jointly for machine reading and showed how to use it to solve span-based question answering tasks. 
We found that \us was able to achieve near state-of-the-art performance on various closed-domain problems such as SQuAD or bAbI, while significantly outperforming the previous state of the art on the open-domain setting on various datasets.

Future work may entail improving the core of the co-encoder. Its architecture is based on BiLSTM building blocks. While they have very powerful sequence modeling capabilities, \cite{dauphin2016language} has shown that they can be replaced by convolutional approaches in the domain of machine translation with both a gain in efficiency and task performance.
Another angle of attack would be to learn jointly parts of the document retriever with the co-encoder in order to maximize the performance of the whole pipeline, as it was attempted in \cite{wang2017r} through reinforcement learning.

\textbf{Acknowledgments} We would like to thank Pranav Rajpurkar for results on the SQuAD test set, Johannes Welbl for results on the WikiHop test set, as well as Louis Martin and Fr\'ed\'eric Arnault for helpful discussions.

\bibliography{weaver}
\bibliographystyle{icml2018}

\end{document}